\documentclass{llncs}

\usepackage{graphicx}
\usepackage{float}
\usepackage{subfig}
\graphicspath{{./figures/}}

\usepackage{amsmath}
\usepackage{amsfonts}
\usepackage{amssymb}
\usepackage{bm}
\usepackage{mathtools}

\usepackage{tikz}
\usetikzlibrary{bayesnet}

\usepackage{minitoc} 
\usepackage[toc,page]{appendix} 
\usepackage[font=small,labelfont=bf]{caption} 
\usepackage{soul}
\usepackage{esvect}
\usepackage{algorithm}
\usepackage{algorithmic}
\usepackage{pgfgantt}
\usepackage{hyperref}
\usepackage{makeidx}
\usepackage[misc]{ifsym}

\DeclareMathOperator*{\argmin}{\arg\!\min} 
\DeclareMathOperator*{\argmax}{\arg\!\max} 

\usepackage{widetable}
\usepackage{booktabs}

\begin{document}

\frontmatter

\pagestyle{headings}

\mainmatter

\title{Nonlinear Markov Random Fields\\Learned via Backpropagation}

\author{Mikael Brudfors \and Ya\"{e}l Balbastre \and John Ashburner}
\institute{The Wellcome Centre for Human Neuroimaging, UCL, London,
UK\\ \email{\{mikael.brudfors.15,y.balbastre,j.ashburner\}@ucl.ac.uk}}


\maketitle

\begin{abstract}
Although convolutional neural networks (CNNs) currently dominate
competitions on image segmentation, for neuroimaging analysis tasks,
more classical generative approaches based on mixture models are still
used in practice to parcellate brains. To bridge the gap between the
two, in this paper we propose a marriage  between a probabilistic
generative model, which has been shown to be robust to variability
among magnetic resonance (MR) images acquired via different imaging
protocols, and a CNN. The link is in the prior distribution over the
unknown tissue classes, which are classically modelled using a Markov
random field. In this work we model the interactions among neighbouring
pixels by a type of recurrent CNN, which can encode more complex spatial
interactions. We validate our proposed model on publicly available MR
data, from different centres, and show that it generalises across
imaging protocols. This result demonstrates a successful and principled
inclusion of a CNN in a generative model, which in turn could be
adapted by any probabilistic generative approach for image segmentation.
\end{abstract}

\section{Introduction}

Image segmentation is the process of assigning one of several
categorical labels to each pixel of an image, which is a fundamental
step in many medical image analyses. Until recently, some of the most
accurate segmentation methods were based on probabilistic mixture
models \cite{klauschen2009evaluation}. These models define a probability
distribution over an observed image ($\vec{X}$), conditioned on
unknown class labels ($\vec{Z}$) and parameters ($\vec{\theta}$).
Assuming a prior distribution over unknown variables, Bayes rule is 
used to form a posterior distribution:
\begin{align}
p(\vec{Z}, \vec{\theta}\mid\vec{X})
\propto p(\vec{X}\mid\vec{Z}, \vec{\theta})~p(\vec{Z},
\vec{\theta}) ~,
\label{eq:bayes-rule}
\end{align}
which can be evaluated or approximated. Wells III \emph{et al.} 
\cite{wells1996adaptive} introduced these types of models for brain
segmentation from magnetic resonance (MR) images. By assuming that the
log-transformed image intensities followed a normal distribution in the
likelihood term $p(\vec{X}\mid\vec{Z},\vec{\theta})$, they segmented
the brain into three classes: grey matter (GM), white matter (WM) and
cerebrospinal fluid (CSF).

As generative models require the data-generating process to be defined, 
they can be extended to more complex joint distributions than in 
\cite{wells1996adaptive}, allowing for segmentation methods robust to, 
\emph{e.g.}, slice thickness, MR contrast, field strength and scanner 
variability. Many of today's most widely used neuroimaging analysis 
software, such as SPM \cite{ashburner2005unified}, FSL 
\cite{zhang2001segmentation} and FreeSurfer \cite{fischl2004sequence}, 
rely on these kinds of models, and have been shown to reliably segment 
a wide variety of MR data  
\cite{kazemi2014quantitative,heinen2016robustness}.

However, recent advances in convolutional neural networks (CNNs) have
provided a new method for very accurate (and fast) image segmentation
\cite{long2015fully}, circumventing the need to define and invert a
potentially complex generative model. Discriminative CNNs learn a
function that maps an input (\emph{e.g.}, an MRI) to an output
(\emph{e.g.}, a segmentation) from training data, where the output is
known. They typically contain many layers, which sequentially apply
convolutions, pooling and nonlinear activation functions to the input
data. Their parameters are optimised by propagating gradients
backwards through the network (\emph{i.e.}, backpropagation). For
medical imaging, the U-net architecture \cite{ronneberger2015u} is the 
most popular and now forms the basis for most top performing entries in 
various medical imaging challenges aimed at segmenting, \emph{e.g.}, 
tumours, the whole brain or white matter 
hyper-intensities\footnote{\url{braintumorsegmentation.org},
\url{wmh.isi.uu.nl}, \url{mrbrains18.isi.uu.nl}}. The more classical
segmentation frameworks based on probabilistic models seem to have met 
their match.
\begin{figure}[t]
\centering
\includegraphics[trim={0cm 0.2cm 0cm
0.2cm},clip,width=0.6\textwidth]{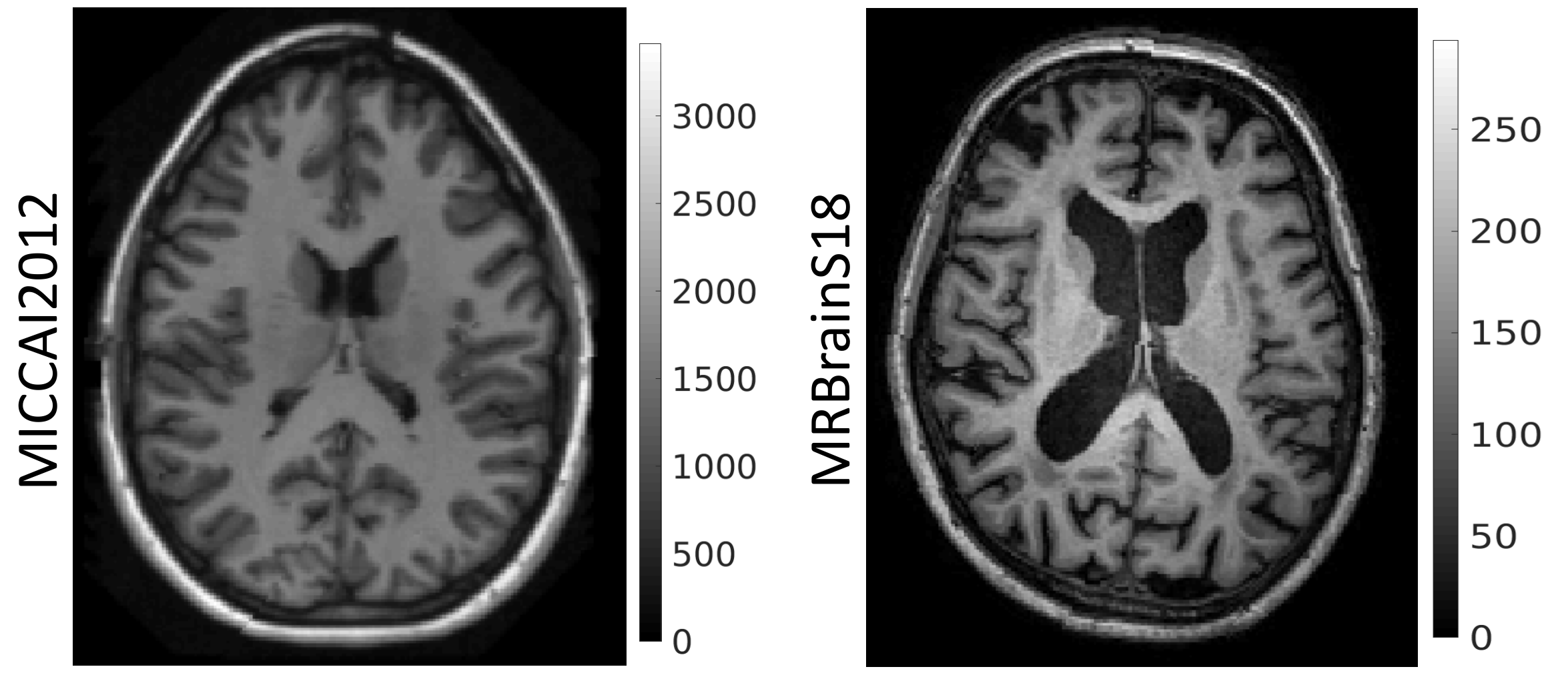}
\caption{T1-weighted MR images from two different, publicly available,
datasets: MICCAI2012 and MRBrainS18 (on which we evaluate our method).
It is evident that learning from one of these populations, and
subsequently testing on the other is very challenging. The intensities
are different by an order of magnitude, the bias is stronger in the
MRBrainS18 subject. Additionally, age related change and pathology can
be clearly seen, such as differences in ventricle size and white matter
hyper-intensities, which further complicates the learning problem.}
\label{fig:compare-t1s}
\end{figure}

Challenges on medical image segmentation can be seen as lab experiments
and -- as with new medical therapies -- there is a large gap to get
\emph{from bench to bedside}. CNNs excel in this context, factorising
the commonalities in an image population of training data, which
generalise to new data from the same population. They can struggle,
however, when faced with new data that contain unseen features 
\cite{dolz20173d}, \emph{e.g.}, a different contrast (Fig. 
\ref{fig:compare-t1s}). This scenario usually requires the model being 
trained anew, on that unseen image contrast. In fact, even without 
considering inter-individual variability (age, brain shape, pathology, 
etc), a CNN-based segmentation software has yet to be presented that is 
agnostic to the great variability in MR data \cite{akkus2017deep}. Lack 
of such software is largely due to the limited amount of labelled data 
available in medical imaging, which is a clear obstacle to their 
generalisability. Some methods have been developed to address this 
problem, \emph{e.g.}, intensity normalisation \cite{han2007atlas}, 
transfer learning \cite{van2015transfer} and batch normalisation 
\cite{karani2018lifelong}. Still, none of these methods are yet general 
enough to solve the task of segmenting across scanners and protocols.
Recently, approaches based on realistic data augmentation have shown promising results \cite{jog2018pulse,zhao2019data}.

In this paper, we propose an approach to bridge between the
classical, but robust, generative segmentation models and more
recent CNN based methods. The link is in the prior term of 
\eqref{eq:bayes-rule}, where we encode the unknown tissue distribution
as drawn from a Markov random field (MRF). Using an MRF is in itself
nothing new; they have been used successfully for decades in order
to introduce spatial dependencies into generative segmentation models
\cite{van1999automated,agn2015brain}, relaxing the independent voxels 
assumption. Here however, we instead  model and learn the interactions 
among neighbouring pixels by a type of CNN. This allows us to 
parametrise the MRF by a more complex mathematical function than in the 
regular linear case, as well as cover a larger neighbourhood than a 
second-order one. The idea is that learning at the tissue level may 
generalise better than learning directly from the image intensities. We 
validate our approach on two publicly available datasets, acquired in 
different centres, and show favourable results when applying the model 
trained on one of these datasets to the other.

\subsubsection*{Related Work:}

Rather than reviewing the use of MRFs in image segmentation we will here
briefly discuss two fairly recent additions to the computer
vision field \cite{zheng2015conditional,schwing2015fully}, because they
are closely related to the method we present in the subsequent section.
The idea of both these papers is to cast the application and learning
of a conditional random field (CRF) into a CNN framework. A CRF is a 
statistical modeling method that directly defines the posterior
distribution in \eqref{eq:bayes-rule}. To compute the CRF both papers 
apply a mean-field approximation, which they implement in the form of a 
CNN.

In contrast to the works described above, we are interested in
defining the full generative model, whilst keeping the separation 
between likelihood and prior in \eqref{eq:bayes-rule}. Modelling these 
two components separately allows us to include expert knowledge and
image-intensity independent prior information over the segmentation
labels. It also integrates easily with existing mixture-model-based
approaches. Furthermore, modelling the prior as an MRF, without
data-dependency in the neighbourhood model, may help in generalising
among different image populations. Finally, our model allows an
arbitrarily complex MRF distribution to be defined, including,
\emph{e.g.}, nonlinearities.

\section{Methods}

In this section we use the generative model defined by 
\eqref{eq:bayes-rule} to encode an MRF over the unknown labels. We show 
that computing this MRF term is analogous to the mathematical 
operations performed by a CNN. We then go on to formulate learning the 
MRF clique potentials as the training of a CNN. This allows us to 
introduce nonlinearities and increasing complexity in the MRF 
neighbourhood.

\subsubsection{Generative Model:}

The posterior in \eqref{eq:bayes-rule} allows us to estimate the 
unknown tissue labels. For simplicity, we will from now on assume that 
all parameters ($\vec{\theta}$) are known; we therefore only want to 
infer the posterior distribution over categorical labels $\vec{Z} \in 
\left\{0,1\right\}^{I \times K}$, where $I$ are the number of pixels in 
the image and $K$ are the number of classes, conditioned on an observed 
image $\vec{X} \in \mathbb{R}^{I\times C}$, where $C$ are the number of 
channels. Modelling multi-channel images allows the use of all acquired 
MR contrasts of the same subject. In practice, unknown parameters of, 
\emph{e.g.}, class-wise intensity distributions would need inferring 
too. Variational Bayesian (VB) inference, along with a well-chosen 
mean-field approximation, allows any such model to fit within the 
presented framework \cite{bishop2006pattern}.

Making use of the product rule, we may define the joint model
likelihood $p(\vec{X},\vec{Z})$ as the product of a data likelihood $p(
\vec{X}\mid\vec{Z})$ and a prior $p(\vec{Z})$. In a mixture model,
it is assumed that once labels are known, intensities are independent
across pixels and all pixels with the same label $k$ are sampled
from the same distribution $p_k(\vec{x})$. This can be written as:
\begin{align}
p(\vec{X}\mid\vec{Z}) = \prod_{i=1}^I\prod_{k=1}^K
p_k(\vec{x}_i)^{z_
{ik}} ~.
\end{align}

A common prior distribution for labels in a mixture model is the
categorical distribution, which can be stationary ($p(\vec{z}_i) =
\mathrm{Cat}\left(\vec{z}_i\mid \vec{\pi}\right)$) or non-stationary
($p(\vec{z}_i) = \mathrm{Cat}\left(\vec{z}_i\mid \vec{\pi}_i\right)$)
\cite{ashburner1997multimodal}. However, both these distributions
assume conditional independence between pixels. MRFs can be introduced
to model dependencies between pixels in a relatively tractable way by 
assuming that interactions are restricted to a finite neighbourhood:
\begin{align}
p(\vec{z}_i\mid
\left\{\vec{z}_j\right\}_{j\neq i}) = p(\vec{z}_i\mid
\vec{z}_{\mathcal{N}_i}) ~,
\end{align}
where $\mathcal{N}_i$ defines pixels whose cliques contain
$\vec{z}_i$. We make the common assumption that this neighbourhood is
stationary, meaning that it is defined by relative positions with
respect to $i$: $\mathcal{N}_i = \left\{i+\delta | \delta \in
\mathcal{N}\right\}$. Here, we assume that this conditional likelihood
factorises over the neighbours and that each factor is a categorical
distribution:
\begin{align}
p(\vec{z}_i\mid\vec{z}_{\mathcal{N}_i})
= \prod_{\delta \in \mathcal{N}} \prod_{k=1}^K \prod_{l=1}^K
\left(\pi_{k,l,\delta}\right)^{z_{i,k} \cdot z_{i+\delta,l}} ~.
\end{align}

\subsubsection*{Mean-field inference:} 

Despite the use of a relatively simple interaction model, the posterior 
distribution over labels is intractable. Therefore, our approach is to 
search for an approximate posterior distribution that factorises across 
voxels:
\begin{align}
p(\vec{Z}\mid\vec{X}) \approx q(\vec{Z}) = \prod_{i=1}^I
q(\vec{z}_i) ~.
\end{align}
We use VB inference \cite{bishop2006pattern}
to iteratively find the approximate posterior $q$ that minimises its
Kullback-Leibler divergence with the true posterior distribution.
Let us assume a current approximate posterior distribution 
$q(\vec{Z})~=~\prod_j\mathrm{Cat}\left(\vec{z}_j\mid\vec{r}_j\right)$; 
each voxel follows a categorical distribution parameterised by 
$\vec{r}_j$, which is often called a \emph{responsibility}. VB then 
gives us the optimal updated distribution for factor $i$ by taking the 
expected value of the joint model log-likelihood, with respect to all 
other variables:
\begin{align}
\ln q^\star(\vec{z}_i) = \sum_{k=1}^K z_{ik} \left(\ln p_k(\vec{x}_i) +
\sum_{\delta\in\mathcal{N}}
\sum_{l=1}^Kr_{i+\delta,l}\ln\pi_{k,l,\delta} \right) + \mathrm{const}.
\end{align}
This distribution is again categorical with parameters:
\begin{align}
r_{ik}^\star \propto \exp\left(\ln p_k(\vec{x}_i) +
\sum_{\delta\in\mathcal{N}} \sum_{l=1}^K r_{i+\delta,l}
\ln\pi_{k,l,\delta}\right).
\end{align}

\subsubsection*{Implementation as a CNN:} 

Under VB assumptions, posterior distributions should be updated one at 
a time, in turn. Taking advantage of the limited support of the 
neighbourhood, an efficient update scheme can be implemented by 
updating at once all pixels that do not share a 
neighbourhood\footnote{When $\mathcal{N}$ contains four second-order 
neighbours, this corresponds to a checkerboard update scheme.}. Another 
scheme can be to update all pixels at once based on the previous state 
of the entire field. Drawing a parallel with linear systems, this is 
comparable to Jacobi's method, while updating in turn is comparable to 
the Gauss-Siedel method. In the Jacobi case, updating the labels'  
expected values can be implemented as a convolution, an addition and a 
softmax operation; three basic layers of CNNs:
\begin{align}
\vec{R}^\star = f(\vec{R}) = \mathrm{softmax}(\vec{C} + \vec{W} \ast
\vec{R}) ~.
\label{eq:mrf-to-cnn}
\end{align}
The matrix $\vec{C}$ contains the conditional log-likelihood terms
($\ln p_k(\vec{x}_i)$). The convolution weights $\vec{W} \in \mathbb{R}^
{\left|\mathcal{N}\right| \times K \times K}$ are equal to the log of
the MRF weights ($\ln\pi_{k,l,\delta}$) and, very importantly, their
centre is always zero. We call such filters \emph{MRF filters}, and the
combination of softmaxing and convolving an \emph{MRF layer}. These 
weights are parameters of the approximate posterior distribution 
$q^\star(\vec{Z})$. Note that multiple mean-field updates can be 
implemented by making the MRF layer recurrent, where the output is 
also the input \cite{zheng2015conditional}. 

Now, let us assume that we have a set of true segmentations
$\vec{\hat{Z}}_{1\dots N}$, along with a set of approximate
distributions with parameters $\vec{R}_{1\dots N}$. One may want to
know the MRF parameters $\vec{W}$ that make the new posterior estimate
$q^\star$ with parameters $\vec{R}^\star = f(\vec{R})$ the most likely
to have generated the true segmentations. This reduces to the
optimisation problem:
\begin{align}
\vec{W}^\star = \argmax_{\vec{W}} \sum_{n=1}^N \ln q^\star(\vec{
\hat{Z}}_n \mid \vec{W}) = \argmax_{\vec{W}} \sum_{i=1}^I \sum_{k=1}^K
\hat{z}_{nik} \ln r^\star_{nik} ~,
\end{align}
which is a maximum-likelihood (ML), or risk-minimisation, problem. Note
that this objective function is the negative of what is commonly
referred to as the categorical cross-entropy loss function in
machine-learning. If the optimisation is performed by computing
gradients from a subset of random samples, this is equivalent to
optimising a CNN, with only one layer, by stochastic gradient descent.

\subsubsection*{Post-processing MRFs:} 

MRFs are sometimes used to post-process segmentations, rather than as 
an explicit prior in a generative model. In this case, the conditional 
data term is not known, and the objective is slightly different: 
approximating a factorised label distribution $q(\vec{Z}) = 
\prod_{i=1}^I q(\vec{z}_i)$ that resembles the prior distribution 
$p(\vec{Z})$. This can be written as finding such distribution $q$
that minimises the Kullback-Leibler divergence with the prior $p$:
\begin{align}
q^\star = \argmin_q \mathrm{KL}\left(q \middle\| p\right) ~.
\end{align}
Again, assuming all other factors fixed with $q(
\vec{z}_j)~=~
\mathrm{Cat}\left(\vec{z}_j\mid\vec{r}_j\right)$, the optimal
distribution for factor $i$ is obtained by taking the expected value of
the prior log-likelihood:
\begin{align}
\ln q^\star(\vec{z}_i) = \sum_{k=1}^K z_{ik} \left(
\sum_{\delta\in\mathcal{N}}
\sum_{l=1}^Kr_{i+\delta,l}\ln\pi_{k,l,\delta} \right) + \mathrm{const},
\end{align}
which is equivalent to dropping the conditional term in the generative
case. Equation \eqref{eq:mrf-to-cnn} is then written as $\vec{R}^\star
= \mathrm{softmax}\left(\vec{W} \ast \Vec{R}\right)$.

\subsubsection*{Nonlinear MRF:} 

\begin{figure*}[t]
\centering
\includegraphics[trim={0cm 0.0cm 0cm 0.1cm},clip,width=\textwidth]{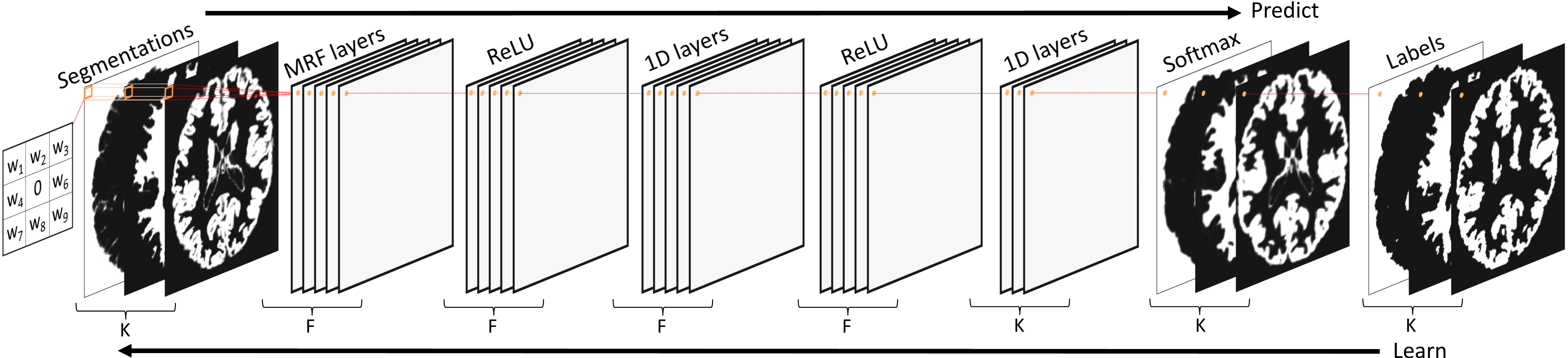}
\caption{An illustration of the architecture of our MRF CNN. Outlined
are the operations performed for learning to predict the centre of a
segmented pixel. The nonlinearities are introduced by the ReLU
activations. By setting the number of MRF layers to $K$ and keeping
only the final softmax layer, the linear MRF model is obtained. The
convolution kernel applied by the MRF filter is shown left of the
segmentations, with its centre constrained to be zero.}
\label{fig:net}
\end{figure*}

The conditional prior distribution 
$p(\vec{z}_i\mid\vec{z}_{\mathcal{N}_i})$ that defines an MRF can, in
theory, be any strictly positive probability distribution. However, in
practice, they are usually restricted to simple log-linear functions,
which are easy to implement and efficient to compute. On the other
hand, deep neural networks allow highly nonlinear functions to be
implemented and computed efficiently. Therefore, we propose a more
complex layer based on multiple MRF filters and nonlinearities, that
implements a nonlinear MRF density. To ensure that we implement a
conditional probability, a constraint is that the input value of a
voxel may not be used to compute its posterior density. Therefore, the
first layer consists MRF filters that do not have a central weight,
and subsequent layers are of size one to avoid reintroducing the centre
value by deconvolution. We thus propose the first layer to be an MRF  
filter $\vec{W} \in \mathbb{R}^{\left| \mathcal{N}\right| \times K 
\times F}$, where $F$ is the number of output features. Setting $F > K$ 
allows the information to be decoupled into more than the initial $K$ 
classes and may help to capture more complex interactions. This first 
convolutional layer is followed by a ReLU activation function, 1D 
convolutions that keep the number of features untouched, and another 
ReLU activation function. This allows features to be combined  
together. A final 1D linear layer is used to recombine  the information 
into $K$ classes, followed by a softmax. Fig. \ref{fig:net} shows our 
proposed architecture.

\subsubsection*{Implementation and training:} 

In this work we set the number of MRF layers to $F=16$, we use three by
three convolutions and leaky ReLU activation functions with 
$\alpha=0.1$. We optimise the CNN using the Adam optimiser. To reduce 
overfitting, we augment the data in two ways: (1) by simple 
left-right reflection; and (2), by sampling warps from anatomically 
feasible affine transformations, followed by nearest neighbour 
interpolation. Realistic affine transformations can be sampled by 
parametrising them by their 12 parameter Lie group (from which the 
transformation matrix can be constructed via an exponential mapping 
\cite{ashburner2013symmetric}) and then learning their mean and
covariance from a large number of subjects' image headers.

\section{Validation}

This section aims to answer a series of questions: (1) does applying a
linear MRF trained by backpropagation to the output segmentations of a
generative model improve the segmentation accuracy? (2) does
complexifying the MRF distribution using numerous filters and
nonlinearities improve the segmentation accuracy compared to a linear
MRF? (3) do the learnt weights generalise to new data from an entirely
different dataset?

\subsubsection{Datasets and preprocessing:}

Our validation was performed on axial 2D slices extracted from two 
publicly available 
datasets\footnote{\url{my.vanderbilt.edu/masi/workshops},
\url{mrbrains18.isi.uu.nl}}:
\begin{itemize}
\item
\textbf{MICCAI2012}: T1-weighted MR scans of 30 subjects aged 18 to 96 
years, (mean: 34, median: 25). The scans were manually segmented into 
136 anatomical regions by Neuromorphometrics Inc. for the MICCAI 2012 
challenge on multi-atlas segmentation.

\item \textbf{MRBrainS18}: Multi-sequence (T1-weighted, T1-weighted 
inversion recovery and T2-FLAIR) MR scans of seven subjects, manually 
segmented into ten anatomical regions. Some subjects have pathology and 
they are all older than 50 years. All scans were labelled by the same 
neuroanatomist.
\end{itemize}
Within each dataset, all subjects were scanned on the same scanner and
with the same sequences, whilst between datasets, the scanners and
sequences differ (Fig. \ref{fig:compare-t1s}). Both datasets have 
multiple labelled brain structures, such as cortical GM, cerebellum, 
ventricles, \emph{etc}. We combined these so as to obtain the same 
three labels for each subject: GM, WM and OTHER ($1 - \text{GM} - 
\text{WM}$). These labels were used as targets when training our model.

\begin{figure*}[t]
\centering
\includegraphics[trim={0cm 0.1cm 0cm
0.1cm},clip,width=\textwidth]{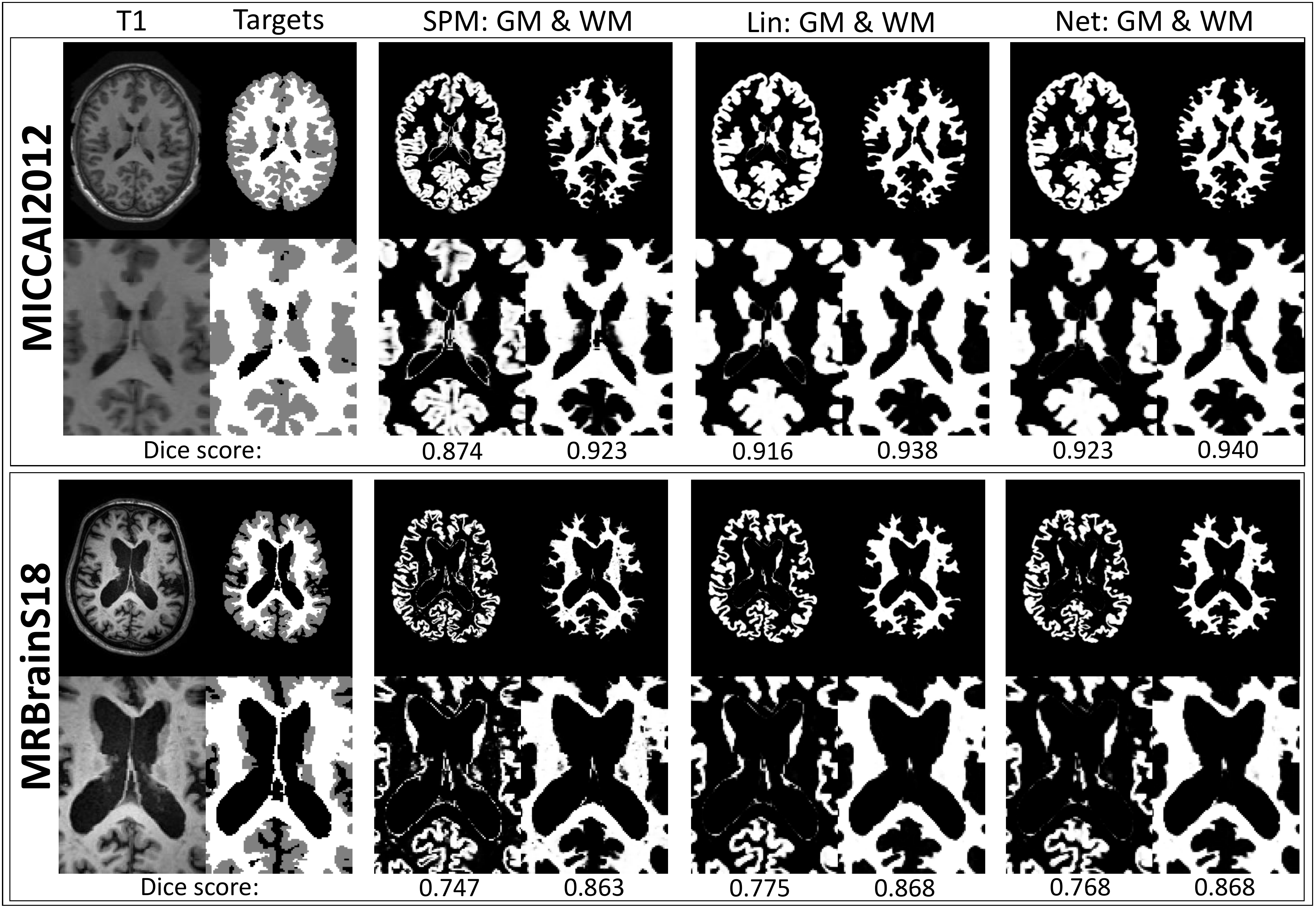}
\caption{Example training data and results. From left to right: 
T1-weighted MR image with target labels, SPM GM and WM segmentations, 
results of applying the linear MRF model to the SPM segmentations, 
results of applying the nonlinear MRF model to the SPM segmentations. 
Below each tissue class are the corresponding Dice scores, computed 
with the target labels as reference.}
\label{fig:results}
\end{figure*}

All T1-weighted MR scans were segmented with the algorithm implemented
in the SPM12
software\footnote{\url{www.fil.ion.ucl.ac.uk/spm/software/spm12}},
which is based on the generative model described in
\cite{ashburner2005unified}. In this model, the distribution over
categorical labels is independent across voxels, non-stationary, and
encoded by a probabilistic atlas deformed towards each subject. The 
algorithm generates soft segmentations, that is, parameters of the
posterior categorical distribution over labels. We pulled the GM, WM 
and OTHER classes from these segmentations. Fig. \ref{fig:results} 
shows the T1-weighted image of one subject from each dataset, with its 
corresponding target labels and SPM12 segmentations\footnote{Besides 
disabling the final MRF clean-up, we used the default parameters of 
SPM12.}.

\subsubsection{Model training and evaluation:}

We trained two different models: a regular, second-order MRF
(\emph{Lin}); and a second-order nonlinear MRF (\emph{Net}). Fig.
\ref{fig:net} explains the differences in architecture between the two.
For each subject and each class, we computed the Dice score of the ML 
labels obtained using SPM12 and those obtained after application of the 
linear MRF and the nonlinear MRF. Statistical significance of the 
observed changes was tested using two-sided Welch's \emph{t}-tests 
between paired measures. Multiple comparisons were accounted for by 
applying the Bonferroni correction.

\begin{figure*}[t]
\centering
\subfloat[]{\includegraphics[trim={0cm 0.7cm 6.8cm
0.2cm},clip,scale=0.8]
{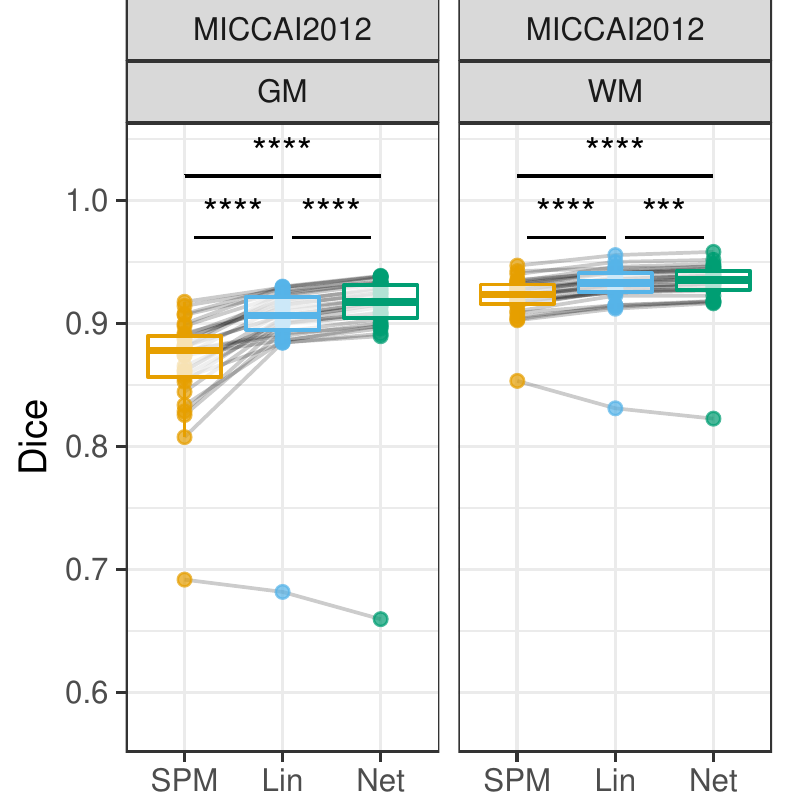}\label{fig:dice:miccai2012}} \hspace{0.25cm}
\subfloat[]{\includegraphics[trim={8.0cm 0.7cm 0cm
0.2cm},clip,scale=0.8]
{dice-all}\label{fig:dice:mrbrains18}}
\caption{Validation of our model on the MICCAI2012 (a) and MRBrainS18 
(b) datasets. Dice scores were computed for known labels (GM and WM) 
and: SPM12 segmentations (SPM); and linear (Lin) and nonlinear (Net)
8-neighbour MRF applied to the SPM12 segmentations. Asterisks indicate statistical significance of paired \emph{t}-tests after Bonferroni
correction: 0.05 ($\ast$), 0.01 ($\ast\ast$), 0.001 ($\ast\ast\ast$) \& 0.0001 ($\ast\ast\ast\ast$).}
\label{fig:dice}
\end{figure*}

We first evaluated the learning abilities of the networks. To this end,
we performed a 10-fold cross validation of the MICCAI2012 dataset, where
groups of three images were tested using a model trained on the 
remaining 27 images. This yielded Dice scores for the entire MICCAI2012 
dataset, which are shown in Fig. \ref{fig:dice:miccai2012}. Next, we 
evaluated the generalisability of the networks, that is, what kind of 
performances are obtained when the models are tested on images from an 
entirely new dataset, with different imaging features. We  randomly 
selected models trained on one of the MICCAI2012 folds and applied them 
to the images from the MRBrainS18 dataset. The results are shown in  
Fig. \ref{fig:dice:mrbrains18}.

\subsubsection{Results:}

The 10-fold cross validation results in Fig. \ref{fig:dice:miccai2012}
show that the increase in Dice scores for both GM and WM is
statistically significant after applying either of our two MRF
CNN models. (Fig. \ref{fig:results} shows the results for a randomly
selected MICCAI2012 subject). With a mean Dice of $\{ \text{GM} =
0.867, \text{WM} = 0.921 \}$ for SPM12, and $\{ \text{GM} = 0.901,
\text{WM} = 0.929 \}$ and $\{ \text{GM} = 0.909, \text{WM} = 0.931 \}$
after applying the linear and nonlinear MRF, respectively. The
results imply that the classical generative approach of SPM12, which
currently ranks in the top 50 on the MRBrainS13 challenge
website\footnote{\url{mrbrains13.isi.uu.nl/results.php}}, could
move up quite a few positions by application of our proposed model
trained on the challenge data. As can be seen in Fig.
\ref{fig:dice:miccai2012}, for one of the subjects, all models perform
substantially worse. On closer inspection, this subject suffers from
major white matter hyper-intensities. This abnormality is currently not
handled well by the MRF CNN models, which obtain lower Dice scores than
the initial SPM12 segmentations.

Fig. \ref{fig:dice:mrbrains18} shows results when applying the models
to data from a different centre, not part of the training data (Fig.
\ref{fig:results} shows the results for a randomly selected MRBrainS18
subject). Mean Dice scores are $\{ \text{GM} = 0.722, \text{WM} = 0.816
\}$ for SPM12, and $\{ \text{GM} = 0.761, \text{WM} = 0.831 \}$ and $\{
\text{GM} = 0.755, \text{WM} = 0.829 \}$ after application of the
linear and nonlinear MRF, respectively. Application of the MRF improves 
both GM and WM segmentations. The nonlinear MRF performs slightly worse 
than the linear version. This result could be due to the nonlinear 
model -- which possesses many more parameters than the linear model -- 
overfitting to the training subjects of MICCAI2012. Additionally, the 
nonlinear MRF may struggle with the MRBrainS18 subjects that have 
pathology (\emph{e.g.} white matter hyperintensities). Still, the fact 
that Dice scores improve when applying the model to new data shows that 
we can successfully improve segmenting images from different MR imaging 
protocols.

\section{Discussion}

In this paper, we introduced an image segmentation method that combines
the robustness of a well-tuned generative model with some of the
outstanding learning capability of a CNN. The CNN encodes an MRF
in the prior term over the unknown labels. We evaluated the method
on annotated MR images and showed that a trained model
can be deployed on an unseen image population, with very different
characteristics from the training population. We hope
that the idea presented in this paper introduces to the medical
imaging community a principled way of bringing together probabilistic
modelling and deep learning.

In medical image analysis -- where labelled training data is sparse and
images can vary widely -- generalisability across different image
populations is one of the most important properties of learning-based 
methods. However, achieving this generalisability is made difficult by 
the limited amount of annotated data; the datasets we used in this 
paper contained, in total, only 37 subjects. This issue may be 
addressed by realistic, nonlinear data augmentation, which is able to 
capture changes due to ageing and disease. Learning this variability in 
shape from a large and diverse population could be a step in that 
direction \cite{balbastre2018diffeomorphic}. On the other hand -- 
manual segmentations suffer from both intra- and inter-operator 
variability, is it clinically meaningful to learn from these very 
imperfect annotations? Could automatic segmentations prove more  
anatomically informative than manual ones (\emph{c.f.}, Fig. 
\ref{fig:results})?  Semi-supervised techniques, leveraging both 
labelled and unlabelled data, could be an option for making 
our method less dependent on annotations (see \emph{e.g.}, 
\cite{roy2018quicknat}).

We chose the architecture of our proposed MRF CNN with the idea of
keeping the number of parameters low (to reduce overfitting), while
still introducing a more complex neighbourhood than in regular MRF
models. However, we did not extensively investigate different
architectures, \emph{e.g.}, activation functions and filter size. There
is therefore a possibility of improved performance by design changes to
the network. Such a change could be to hierarchically apply MRF filters
of decreasing size, which could increase neighbourhood size without
increased overfitting. Another potentially interesting  idea would be
to `plug in' the MRF filters at the end of a segmentation network, such
as a U-net, emulating MRF post-processing inside the network. Finally,
we intend to integrate our model into a generative segmentation
framework and then  validate its performance by comparing it to other
existing segmentation software.


\subsubsection*{Acknowledgements:} JA was funded by the EU Human Brain
Project's Grant Agreement No 785907 (SGA2). YB was funded by the MRC
and Spinal Research Charity through the ERA-NET Neuron joint call
(MR/R000050/1).

\bibliography{bibliography}%
\bibliographystyle{ieeetr}%
\end{document}